\documentclass{jors}

\usepackage{natbib}
\usepackage{microtype}
\usepackage{graphicx}
\usepackage{subfigure}
\usepackage{amsmath}
\usepackage{booktabs} 

\definecolor{mygray}{gray}{0.6}

\begin{document}

\rule{\textwidth}{1pt}

\section*{(1) Overview}

\vspace{0.5cm}

\section*{Title}

Recent Advances of NeuroDiffEq -- An Open-Source Library for Physics-Informed Neural Networks

\section*{Paper Authors}

Liu, Shuheng

Protopapas, Pavlos

Sondak, David

Chen, Feiyu

\section*{Paper Author Roles and Affiliations}
Shuheng Liu: Ideation and Development, Corresponding Author, John A. Paulson School of Engineering and Applied Science, Harvard University

Pavlos Protopapas: Ideation, John A. Paulson School of Engineering and Applied Science, Harvard University

David Sondak: Ideation, John A. Paulson School of Engineering and Applied Science, Harvard University

Feiyu Chen: Ideation and Development, John A. Paulson School of Engineering and Applied Science, Harvard University

\section*{Abstract}

Solving differential equations is a critical challenge across a host of domains. While many software packages efficiently solve these equations using classical numerical approaches, there has been less effort in developing a library for researchers interested in solving such systems using neural networks. With \textit{PyTorch} as its backend, NeuroDiffEq is a software library that exploits neural networks to solve differential equations. In this paper, we highlight the latest features of the NeuroDiffEq library since its debut. We show that NeuroDiffEq can solve complex boundary value problems in arbitrary dimensions, tackle boundary conditions at infinity, and maintain flexibility for dynamic injection at runtime.

\section*{Keywords}

Physics-Informed Neural Networks, Differential Equation, Numerical Methods, Scientific Computing, Physics-Informed Deep Learning

\section*{Introduction}


\paragraph{Overview} 
Differential equations can be found across a host of scientific domains including economics, genetics, physics, engineering, chemistry, and biology. 
Therefore, it is a critical challenge to accurately and efficiently compute their solutions. 
While traditional methods for temporal (e.g. Runge-Kutta methods) and spatial discretization (e.g. finite element methods) have been extensively studied to tackle this problem, more recent work has identified neural networks as potentially more robust and powerful \citep{raissi_physics_2017,raissi_physics-informed_2019}. 
Neural networks offer potential advantages over and enhancements of traditional numerical approaches to solving differential equations. These include: 1.) they can represent complicated functions with a few neurons, 2.) neural networks can be substantially less storage intensive, 3.) they can be used naturally for transfer learning, which, in addition to other capabilities, provides generalizations of classical numerical continuation, and 4.) neural networks can learn a bundle of solutions to a parameterized family of equations. Moreover, neural networks are differentiable functions, so solutions to differential equations learned with neural networks are continuously differential in space and time.
Collectively, these advantages provide a strong basis for learning solutions to differential equations particularly. 
Given this, we developed NeuroDiffEq -- a novel software library that allows researchers and engineers to solve differential equations using neural networks. 
Since its inception \citep{ndiffeq}, the library has grown significantly to encompass additional domains, novel sampling techniques, and a unique corpus (NeuroDiffHub) where users can share their pre-trained differential equations for transfer learning. 
Here, we unpack the latest developments highlighting the contribution, potential application, and usage.
As an overview, new capabilities of NeuroDiffEq include 
\begin{itemize}
    \item Generic solver for high dimensional PDEs,
    \item Bundle solver for ODE systems with parameterized constants (either in initial conditions or equations),
    \item Dynamic behavior injection at runtime,
    \item Built-in support for operators and function bases, and
    \item More reparameterizations and loss functions.
\end{itemize}

\paragraph{Background}
Neural networks are universal function approximators \citep{hornik_multilayer_1989}, which make them a natural tool for solving differential equations. 
Consider a differential equation $\mathcal{D} \mathbf{u} = \mathbf{f}$, where $\mathcal{D}$ is a (possibly nonlinear) differential operator and $f$ is some forcing function.
A network can be trained to approximate the solution $\mathbf{u}$ by minimizing the Monte-Carlo $L_2$ norm of residual $(\mathcal{D} \mathbf{u} - \mathbf{f})$ over the equation domain \citep{lagaris_artificial_1998, raissi_physics_2017}.
Evaluating $\mathcal{D}\mathbf{u}$ involves computing derivatives of various orders using automatic differentiation. 
Furthermore, it has been shown that initial and boundary conditions can be enforced on the solution $\mathbf{u}$ through a reparametrization trick \cite{mattheakis_hamiltonian_2020}. 

With \textit{PyTorch} as a its backend, \href{https://neurodiffeq.readthedocs.io/en/latest/}{NeuroDiffEq} provides a toolbox to solve differential equations using neural networks. 
The library works out of the box for machine learning practitioners while remaining immensely flexible for ad hoc customization. 
All components of the training process can be customized, including network architecture, reparametrization, optimizer, loss function, sampling techniques, and dynamic hyperparameter adjustment during training. 

\paragraph{Related Work}
In their seminal work, Lagaris et al \citep{lagaris_artificial_1998} formulated the idea of solving differential equations using neural networks.
While the results of that paper were promising, they were only brought back when physics informed neural networks (PINNs) were introduced \citep{raissi_physics_2017}. 
PINNs solve differential equations in fundamentally the same way except the gradients of the solution are not computed analytically but numerically via backpropagation. 
In other words, the advent of novel AI techniques and machinery allowed the authors of PINNs to show the true scalability of learning differential equations with neural networks. 
Since then, numerous efforts in this direction have shown that such a framework is particularly powerful at learning across a host of domains \citep{yang_learning_2020,wang_train_2021,sahli_costabal_physics-informed_2020,he_physics-informed_2020}.
The NeuroDiffEq library has been extensively downloaded and used since its inception. As of September 2023, the library has been downloaded 34,000 times from PyPI alone. Using this library, there have also been efforts to establish error bound for physics-formed neural networks on certain classes of differential equations \citep{liu2022evaluating, pmlr-v216-liu23b}.

\section*{Implementation and architecture}


\paragraph{Constructing the Solver}
The NeuroDiffEq library uses PyTorch as its backend for automatic differentiation.
At the core of the NeuroDiffEq library is the \texttt{BaseSolver} class, which, among other configurable options, takes in the following inputs:
\begin{enumerate}
    \item A callable object (e.g. functions) that represents the residual of the differential equation,
    \item A list of \texttt{BaseCondition} objects that represent the initial and/or boundary conditions,
    \item (Optional) Two \texttt{Generator}s, used for generating random collocation points for training and validation, respectively,
    \item (Optional) The neural network architecture, and
    \item (Optional) The optimizer along with its hyperparameters.
\end{enumerate}

\paragraph{Training for the Solution}
Once constructed, the neural networks can be trained by calling the \texttt{fit()} method of the solvers object. 
At this stage, a sequence of callback functions can be specified for dynamic bahavior injection after each training epoch.

\paragraph{Accessing the Solution}
Once the training completes, the solution can be accessed by calling the \texttt{get\_solution()} method of the solver object. 
The solution obtained is a callable Python object (i.e., a function). In particular, the input and output can either be Numpy arrays or PyTorch tensors.

\section*{Quality control}


The library can be tested using \textit{pytest} on Linux, macOS, and Windows. Travis CI is used for continuous integration of the library.
A user can verify the installation by downloading the unit tests from the library's GitHub repository and running \texttt{pytest} in the test directory. As of the time of writing, the coverage of test cases has reached 85\%.

\section*{(2) Availability}
\vspace{0.5cm}
\section*{Operating system}


NeuroDiffEq works on all major operating systems including Linux distributions, macOS, and Windows, as long as relevant dependencies (Python 3.6 or later and PyTorch 1.0 or later) are installed.

\section*{Programming language}


Python 3.6 or later

\section*{Additional system requirements}


NeuroDiffEq can be run on a variety of hardware configurations. To balance speed and compatibilty, it automatically switches to GPU devices if available and defaults back to CPU otherwise. 

The memory and disk space footprint depends primarily on the network architecture, size of mini-batches, and complexity of the target equations. Any reasonable system (such as one with 2 GiB memory and 1 CPU) should be able to run the library. However, scarcity of resources necessitates performing multiple forward passes per mini-batch, which slows down training.

\section*{Dependencies}


\begin{itemize}
	\item pytorch 1.0 or later
	\item dill 0.3.3 or later
	\item tqdm 4.40 or later
	\item tensorboard 2.0 or later (required for visualization only)
	\item matplotlib 3.0 or later (required for visualization only)
	\item seaborn 0.10.0 or later (required for visualization only)
	\item pytest 5.0 or later (required for testing only)
\end{itemize}

\section*{List of contributors}
\begin{enumerate}
    \item Shuheng Liu; lead developer of the library; Harvard John A. Paulson School of Engineering and Applied Sciences, Cambridge, MA, United States.
    \item Feiyu Chen; originator of the library; Harvard John A. Paulson School of Engineering and Applied Sciences, Cambridge, MA, United States.
    \item Shivas Jayaram; contributor of the library;
    \item Joy Parikh; contributor of the library;
    \item David Sondak; contributor and advisor of the library, Harvard John A. Paulson School of Engineering and Applied Sciences, Cambridge, MA, United States.
    \item Lan Bi; contributor of the library; New York University, New York, NY, United States.
    \item Augusto T. Chantada; contributor of the library; Departamento de Física, Facultad de Ciencias Exactas y Naturales, Universidad de Buenos Aires, Buenos Aires, Argentina.
    \item Sakthisree Venkatesan; contributor of the library; Univ.ai, Bangalore, India.
    \item Sathvik Bhagavan; contributor of the library; Indian Institute of Technology, Kanpur, India.
    \item M. Elaine Cunha; contributor of the library; Harvard John A. Paulson School of Engineering and Applied Sciences, Cambridge, MA, United States.
    \item Devansh Agarwal; contributor of the library; Department of Physics and Astronomy, Virginia University, Morgantown, WV, United States; Center for Gravitational Waves and Cosmology, West Virginia University, Morgantown, WV, United States.
    \item Matin Moezzi; contributor of the library; University of Toronto, Toronto, ON, Canada.
    \item Jo\~{a}o Esteves; contributor of the library;
    \item Marco Di Giovanni; contributor of the library; Politecnico di Milano, Milano, Italy.
    \item Pavlos Protopapas; advisor of the library; Harvard John A. Paulson School of Engineering and Applied Sciences, Cambridge, MA, United States.
\end{enumerate}

\section*{Software location:}

{\bf Archive} 


\begin{description}[noitemsep,topsep=0pt]
	\item[Name:] GitHub
	\item[Persistent identifier:] \,\\
    https://github.com/NeuroDiffGym/neurodiffeq/archive/refs/tags/v0.6.2.zip
	\item[Licence:] MIT
	\item[Publisher:]  Shuheng Liu
	\item[Version published:] v0.6.2
	\item[Date published:] 08/06/23
\end{description}

{\bf Code repository} GitHub


\begin{description}[noitemsep,topsep=0pt]
	\item[Name:] NeuroDiffEq
	\item[Persistent identifier:] https://github.com/NeuroDiffGym/neurodiffeq
	\item[Licence:] MIT License
	\item[Date published:] 24/03/19
\end{description}



\section*{Language}


Python 3

\section*{(3) Reuse potential}


Many new features have been introduced to NeuroDiffEq since its debut in 2020.
The library is supported by its developers via GitHub issues and pull requests.
The URL of GitHub repository is listed above in the software location section.

The use cases of this library are described in detail below.

\begin{figure*}[ht]
    \centering
    \includegraphics[width=.85\textwidth]{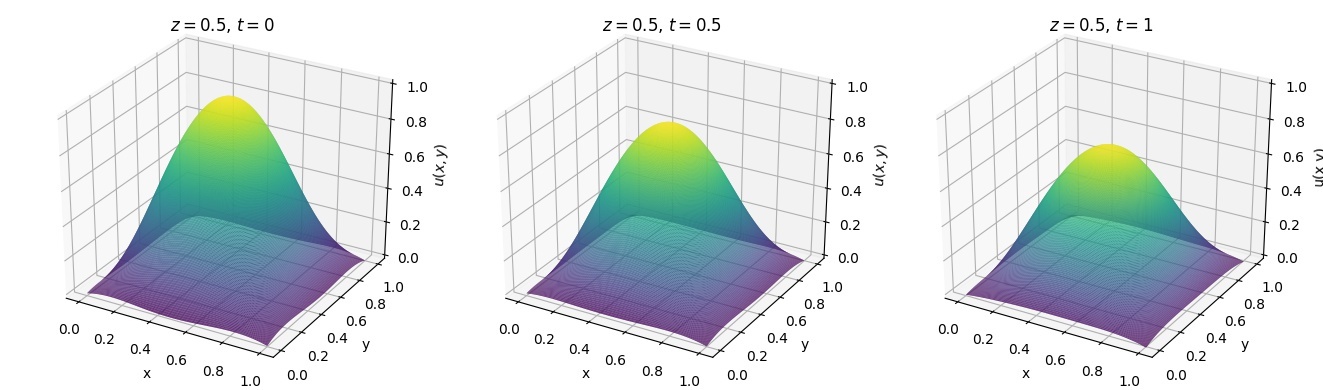}
    \caption{Time evolution of the 3-dimensional heat equation at $z=0.5$}
    \label{fig:heateq-surf}
\end{figure*}

\begin{figure*}[ht]
    \centering
    \includegraphics[width=.85\textwidth]{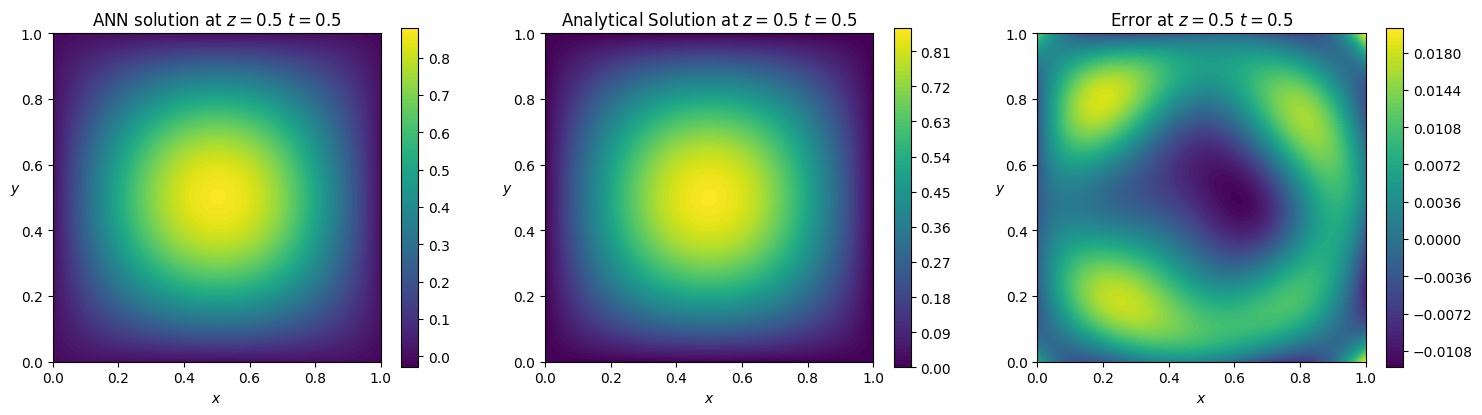}
    \caption{Comparison of ANN solution and True Solution}
    \label{fig:heateq-contourf}
\end{figure*}

\paragraph*{Generic Solver and Composable Generators}
While previous versions of NeuroDiffEq were only capable of solving ODEs and low-dimensional PDEs ($<2$ dimensions), the new library is capable of solving these systems for arbitrary dimensions. To demonstrate, we solve $D$-dimensional heat equations on the hypercube $\Omega = [0, 1]^D$ for $t \in [0, 1]$ where $D = 1, 2, \dots, 10$.
\begin{equation}
    \frac{\partial u}{\partial t} = \frac{1}{100} \nabla_{\mathbf{x}} u \quad \text{for} \quad (t, \mathbf{x}) \in [0, 1] \times [0, 1]^D,
\end{equation}
under the initial condition $u|_{t=0} = \prod_{d=1}^{D} \sin(\pi x_d)$ and the boundary condition $u|_{\partial \Omega} = 0$.
For illustration purposes, we plot the solution over time for $D=3$ in Fig.~\ref{fig:heateq-surf}. The inputs are $(x, y, z, t)$ and we visualize the $(x, y)$-plane slice at $z=0.5$ and $t\in \left\{0, 1\right\}$. Also, we visualize the error against the analytical solution at $z=0.5$ and $t=0.5$ in Fig.~\ref{fig:heateq-contourf}. The code to generate this example can be found \href{https://colab.research.google.com/drive/1lPdJwSP0ziVMxR2yDppbkJqOKtALQ1id?usp=sharing}{here}.


In addition to generalized higher-order solvers, the library also extends to customizable generators. Generators, as the name suggests, generate random collocation points to form the training domain and can now be composed to represent more complex geometries in higher dimensions.

\begin{figure}[ht]
    \centering
    \includegraphics[width=0.4\textwidth]{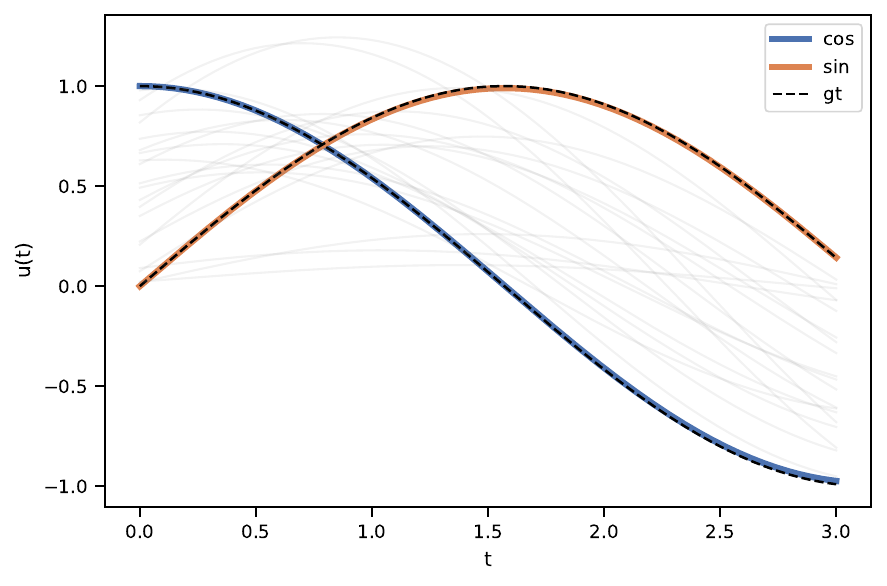}
    \includegraphics[width=0.4\textwidth]{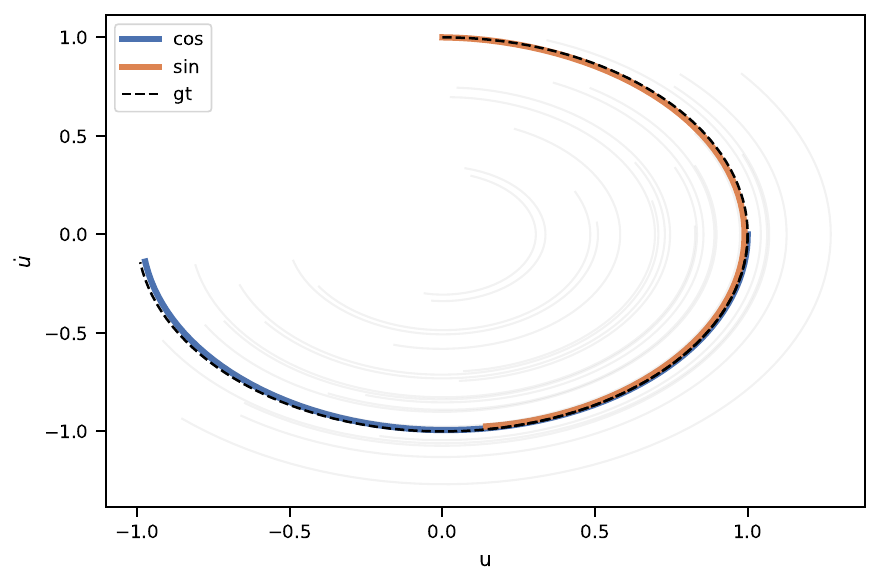}
    \caption{Bundle solvers are trained on multiple initial conditions simultaneously and are therefore robust at inference to new initial conditions.}
    \label{fig:bundles}
\end{figure}

\paragraph*{Bundle Solver for ODE Systems}
For ODE systems with parametrizable equation setup and initial conditions, the library offers a BundleSolver1D that solves for a bundle of solutions $\mathbf{u}(t; \Theta_{\text{ic}}, \Theta_{\text{eq}})$, where $\Theta_{\text{ic}}$ is variable parameters for the initial condition and $\Theta_{\text{eq}}$ is variable parameters for the equation. This approach is more efficient than solving for one set of fixed parameters at a time \citep{flamant_solving_2020}. An example to showcase this performance is the simple harmonic oscillator which exhibits sine and cosine solutions depending on the initial conditions. We consider a set of initial conditions $\left\{u_{k}(t_{0}), \dot{u}_{k}(t_{0})\right\}_{k=1}^{N_{\text{ic}}}$ for the system of equations with $t_{0}=0$ such that $u_{k}(t_{0}), \dot{u}_{k}(t_{0}) \in [0, 1]$.  We find that with 5000 iterations the bundle solver is able to identify the correct trajectories for initial conditions that lead to both sine and cosine solutions (see Fig. \ref{fig:bundles}).

As another example, consider a case where a user wants to simultaneously solve a bundle of differential equations of the form $\displaystyle \frac{\mathrm{d}u}{\mathrm{d}t} + \lambda u = 0$, under the initial condition $u(0) = u_0$. By treating $\lambda$ and $u_0$ as inputs to the neural network, it is possible to solve the inverse problem for $(\lambda, u_0)$.

This approach proves beneficial for understanding processes like chemical reactions with unknown rates. Here, distinct reaction rates corresponds to different solutions, and only one aligns with observed data. The user can solve for a bundle of solutions first, then optimize the reaction rates. This process is known as the inverse problem.

After training the model, the user can derive the solution expected in $(t, \lambda, u_0)$ inputs. The user can then leverage simple gradient descent to match the bundled solutions against observed data points $(t_i, u_i)$, leading to the identification of best parameters $(\lambda, u_0)$.

This model not only provides a simultaneous solution of equations but also enables the resolution of inverse problems, like determining unknown parameters in the equation.

\paragraph*{Dynamic Behavior Injection at Runtime}
The library allows users to perform certain actions based on customized conditions while training using the callback feature. A callback is a function that specifies a sequence of action to be performed on the solver after each training epoch. As the training progresses, users can perform any actions, such as
\begin{itemize}
    \item changing the loss function, optimizer, batch size, problem domain, sampling strategy, etc.;
    \item plotting visualizations, write tensorboard summaries, print logs, or save checkpoints to disk;
    \item freezing certain layers or performing early stopping.
\end{itemize}

To specify if and when an action should be carried out, users can specify conditions, which are composable with boolean operators \textbf{and} (\&), \textbf{or} (\texttt{|}), \textbf{not} ($\sim$), and \textbf{xor} ($\wedge$). These conditions are checked after training and validation in every epoch. An action will be performed if and only if its associated condition is met.

In the example below, we train initially with $L_2$ loss and change to $L_1$ once the validation $L_2$ loss converges ($\Delta < 10^{-4}$) for 20 consecutive epochs. We train the network for a total of 1000 epochs. This strategy is easy to implement in NeuroDiffEq and outperforms training consistently with either $L_1$ or $L_2$ by themselves.

\begin{figure}[ht]
    \centering
    \includegraphics[width=0.5\textwidth]{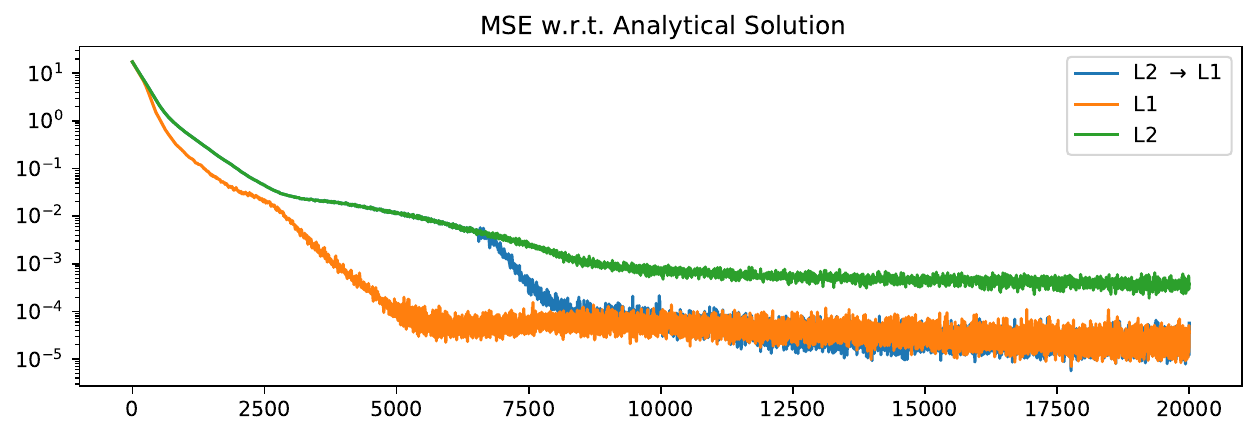}
    \caption{Comparison of MSE against True Solution}
    \label{fig:changing-loss}
\end{figure}

\paragraph*{Operator and Function Basis}
Recent versions of NeuroDiffEq include more effcient operator implementations. In older versions, the implementation of the gradient operator required as many backward passes as there were independent variables in the problem (e.g. three backward passes for a system with three spatial dimensions). New versions of NeuroDiffEq include gradient ($\nabla$), divergence ($\nabla\cdot$), curl ($\nabla\times$), scalar Laplacian ($\nabla^2$), and vector laplacian ($\Delta$) in Cartesian, spherical, and cylindrical coordinate systems. These library operators give up to a $3\times$ boost in speed and consume less GPU memory by computing multiple partial derivatives per backward pass and reusing computational graphs where possible. See Table \ref{table:operator-speed} for performance comparison.

\begin{figure*}[ht]
    \centering
    \includegraphics[width=\textwidth]{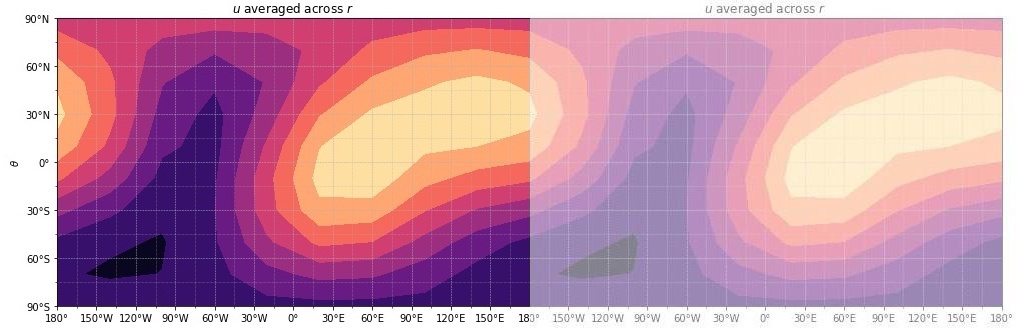}
    \caption{An equirectangular projection of the solution to Eq. \ref{eq:gaussian-charge} averaged across radius $r$. The solution is concatenated with a copy of itself to the right, showing that the solution is periodic in the azimuth.}
    \label{fig:pbc}
\end{figure*}

We also support expanding solutions using function bases including real Fourier series, real spherical harmonics, and zonal harmonics to satisfy periodic boundary conditions. To illustrate, we solve a Poisson equation w.r.t. electric potential of a Gaussian charge density 
\begin{equation}\label{eq:gaussian-charge}
    \nabla^2 u(r, \theta, \phi) = \frac{1}{\sqrt{2\pi}^3} \exp\left(-\frac{r^2}{2}\right)
\end{equation} in spherical coordinates. The solution given by the neural network not only converges nicely to the true solution, but also exactly satisfies the periodic boundary conditions, as show in Figure \ref{fig:pbc}.

\begin{table}
\resizebox{\textwidth}{!}{
\begin{tabular}{c|c|c|c|c}
  Coordinate   & Operator & Benchmark (ms) & Optimized (ms) & Speedup (x) \\
\hline 
    Cylindrical & Grad & $1.560 \pm 0.326$ & $0.532 \pm 0.110$ & 2.93 \\
    Cylindrical & Divergence & $1.737 \pm 0.392$ & $1.591 \pm 0.261$ & 1.09\\
    Cylindrical & Laplace & $1.996 \pm 0.391$ & $1.793 \pm 0.262$ & 1.11\\
    Cylindrical & Vector Lap & $13.323 \pm 1.384$ & $8.815 \pm 0.976$ & 1.51\\
    Cylindrical & Curl & $3.241 \pm 0.429$ & $1.592 \pm 0.293$ & 2.04\\
     
     Spherical & Grad & $1.600 \pm 0.286$ & $0.639 \pm 0.138$ & 2.50\\
     Spherical & Divergence & $2.111 \pm 0.317$ & $2.009 \pm 0.699$ & 1.05\\
     Spherical & Laplace & $4.462 \pm 0.550$ & $3.330 \pm 0.642$ & 1.34\\
     Spherical & Vector Lap & $17.299 \pm 1.741$ & $9.946 \pm 1.047$ & 1.74\\
     Spherical & Curl & $3.615 \pm 0.513$ & $1.799 \pm 0.226$ & 2.01\\
     
     Cartesian & Grad & $1.512 \pm 0.450$ & $0.501 \pm 0.137$ & 3.02\\
     Cartesian & Divergence & $1.472 \pm 0.289$ & $1.595 \pm 0.441$ & 0.92\\
     Cartesian & Laplace & $3.808 \pm 0.567$ & $2.862 \pm 0.339$ & 1.33\\
     Cartesian & Vector Lap & $11.367 \pm 1.190$ & $8.523 \pm 1.033$ & 1.33\\
     Cartesian & Curl & $3.080 \pm 0.424$ & $1.529 \pm 0.329$ & 2.01\\
\end{tabular}
}%
\caption{Comparison of Operator Speed}
\label{table:operator-speed}
\end{table}

\paragraph*{More Reparameterizations and Loss Functions}
New reparameterizations have been added, including notably a double-sided Dirichlet condition with one side at infinity: $$u|_{r_0} = u_0 \qquad \lim_{r\to \infty} u(r) = u_\infty.$$ 
We enforce such constraints using the reparameterization
\begin{align}
    u(r) &= \tanh(r-r_0) u_\infty + \exp\left(-(r-r_0)\right) u_0 + \nonumber \\
         &\quad \tanh(r-r_0)\exp\left(-(r-r_0)\right) \mathrm{Net}(r)    
\end{align}
which can be easily generalized to spherical, cylindrical, and polar coordinates by substituting constants $u_0$ and $u_\infty$ with functions independent of $r$ (e.g. $u_0(\theta, \phi)$ and $u_\infty(\theta, \phi)$ for spherical coordinates).
We also implemented mixed Dirichlet and Neumann conditions for a closed interval 
\begin{align*}
    u(x_0) = u_0 \quad &u(x_1) = u_1 \\
    u'(x_0) = u'_0 \quad &u(x_1) = u_1 \\
    u(x_0) = u_0 \quad &u'(x_1) = u'_1 \\
    u'(x_0) = u'_0 \quad &u'(x_1) = u'_1
\end{align*}
using the reparameterization described in \cite{lagari2020systematic}. On top of these, users can customize reparameterization to satisfy other initial or boundary conditions. 

\begin{figure}[ht]
    \centering
    \includegraphics[width=0.45\textwidth]{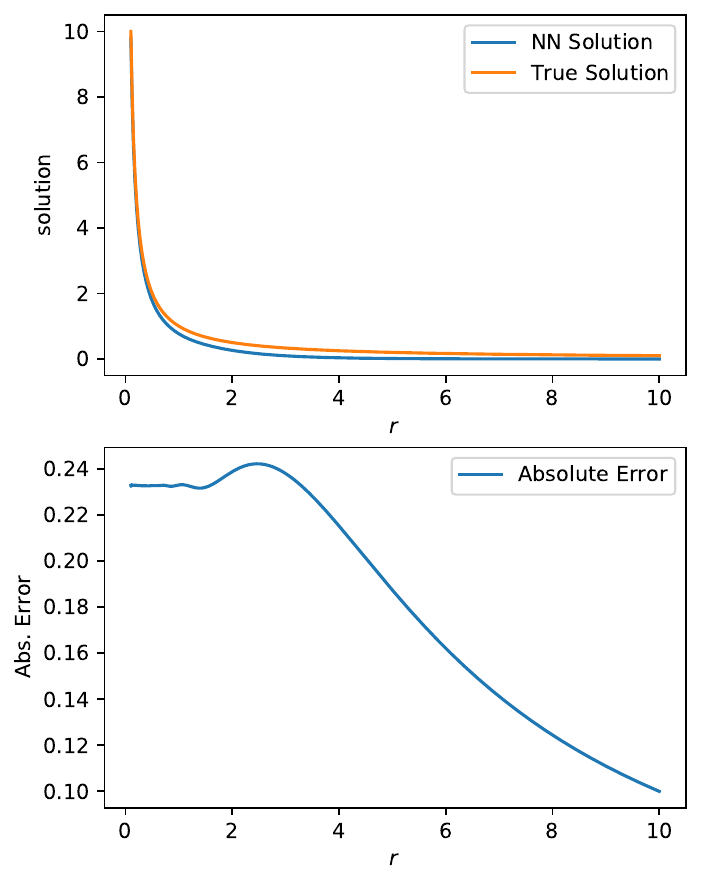}
    \caption{Solution of Gravitational Potential}
    \label{fig:gravitational-potential}
\end{figure}

In Figure \ref{fig:gravitational-potential}, we solve the gravitational potential equation $$\frac{\mathrm{d}E_p}{\mathrm{d}r} = - \frac{GMm}{r^2}\quad \text{s.t.}\quad E_p\big | _{r=\infty}=0.$$ with respect to $E_p(r)$.
In addition to the $L_2$ loss function of ODE/PDE residual $R(\mathbf{x})$
$$
    L_2(\theta) = \left( \int_{\mathbf{x} \in \Omega} \big\|R(\mathbf{x};\theta)\big\| ^2 \mathrm{d}\mathbf{x} \right)^{\frac{1}{2}}, 
$$
there have been more built-in loss functions, including $L_1$, $L_\infty$, $H_1$ and semi-$H_1$. Users can also customize loss functions add include regularization terms for specific needs, including enforcing more complicated boundary conditions. 

\paragraph*{Efficiency}

Due to the reverse mode of automatic differentiation that \textit{PyTorch} employs, we can perform multiple forward and backward passes per optimization step. This effectively allows arbitrarily large batch sizes on computers with limited memory at the cost of speed. NeuroDiffEq supports this technique by default, except for the L-BFGS optimizer, which doesn't permit such operation.

When it is imported, NeuroDiffEq detects whether a compatible GPU device is available, in which case it automatically runs the computation on GPU. 
The library also defaults to 64-bit float for higher numerical precision. 
Users can also manually switch to another device (including the neural engine of Apple M-series chips) or another floating precision.

\section*{Acknowledgements}
The authors appreciates the contribution of Shaan Desai, who helped with the ideation of the organization of this paper.


\section*{Funding statement}

The softeware did not result from funded research.

\section*{Competing interests}


The authors declare that they have no competing interests.

\bibliography{references}

\vspace{2cm}

\rule{\textwidth}{1pt}

{ \bf Copyright Notice} \\
Authors who publish with this journal agree to the following terms: \\

Authors retain copyright and grant the journal right of first publication with the work simultaneously licensed under a  \href{http://creativecommons.org/licenses/by/3.0/}{Creative Commons Attribution License} that allows others to share the work with an acknowledgement of the work's authorship and initial publication in this journal. \\

Authors are able to enter into separate, additional contractual arrangements for the non-exclusive distribution of the journal's published version of the work (e.g., post it to an institutional repository or publish it in a book), with an acknowledgement of its initial publication in this journal. \\

By submitting this paper you agree to the terms of this Copyright Notice, which will apply to this submission if and when it is published by this journal.

\end{document}